\documentclass[a4wide]{article}
\usepackage{booktabs} 
\usepackage[graphicx]{realboxes}

\usepackage[vlined,linesnumbered,ruled,titlenotnumbered]
{algorithm2e}

\usepackage{mathtools}
\usepackage{hyperref}
\usepackage{balance}
\usepackage{xspace}
\usepackage{algpseudocode,booktabs,amsmath}
\usepackage{multirow, rotating}
\usepackage{mwe,subfigure}
\usepackage{color}
\usepackage{blindtext}
\usepackage{amssymb}
\usepackage{rotating}
\usepackage{amssymb}
\usepackage{hyperref}
\usepackage{balance}
\usepackage{tikz}
\usetikzlibrary{shapes,decorations,patterns,positioning}
\usetikzlibrary{plotmarks}
\usepackage{url}
\usepackage{lmodern}
\usepackage{ctable}
\usepackage[english]{babel}

\newcommand{\GCF}{\mbox{\em{GCF}}}
\DeclareSymbolFont{symbolsC}{U}{pxsyc}{m}{n}
\DeclareMathSymbol{\colonequals}{\mathrel}{symbolsC}{"42}

\newcommand{\symm}{\mbox{\em{Symmetry}}}
\newcommand{\hue}{\mbox{\em{Hue}}}
\newcommand{\sdhue}{\mbox{\em{SDHue}}}
\newcommand{\sat}{\mbox{\em{Saturation}}}
\newcommand{\smooth}{\mbox{\em{Smoothness}}}
\usepackage{graphicx}
\usepackage{booktabs}
\usepackage{multirow}
\newcommand{\mlea}{(\mu + \lambda)-EA_D}

\newcommand{\tmax}{\ensuremath{t_{\max}}\xspace}

\newcommand{\ignore}[1]{}
\DeclareMathOperator{\Vol}{Vol}

\title{Discrepancy-based Evolutionary
 Diversity Optimization}

\author{
Aneta Neumann$^1$, Wanru Gao$^1$, Carola Doerr$^2$,\\
Frank Neumann$^1$, Markus Wagner$^1$\\
\ \\
$^1$Optimisation and Logistics, The University of Adelaide,\\ Adelaide, Australia\\
$^2$CNRS and Sorbonne Universit\'es, UPMC Univ Paris 06,\\ LIP6, Paris, France
}

\begin{document}

\maketitle
\sloppy

\begin{abstract}
Diversity plays a crucial role in evolutionary computation. While diversity has been mainly used to prevent the population of an evolutionary algorithm from premature convergence, the use of evolutionary algorithms to obtain a diverse set of solutions has gained increasing attention in recent years. Diversity optimization in terms of features on the underlying problem allows to obtain a better understanding of possible solutions to the problem at hand and can be used for algorithm selection when dealing with combinatorial optimization problems such as the Traveling Salesperson Problem. We explore the use of the star-discrepancy measure to guide the diversity optimization process of an evolutionary algorithm. 

In our experimental investigations, we consider our discrepancy-based diversity optimization approaches for evolving diverse sets of images as well as instances of the Traveling Salesperson problem where a local search is not able to find near optimal solutions. Our experimental investigations comparing three diversity optimization approaches show that a discrepancy-based diversity optimization approach using a tie-breaking rule based on weighted differences to surrounding feature points provides the best results in terms of the star discrepancy measure.
\end{abstract}

\section{Introduction}

Diversity plays a crucial role in evolutionary computation. Traditionally, diversity is used to avoid premature convergence and it is generally assumed that crossover-based evolutionary algorithms need a diverse population in order to produce good results.

During the last 10 years, using evolutionary algorithms to produce a diverse set of solutions has gained increasing attention. Ulrich and Thiele~\cite{DBLP:conf/gecco/UlrichT11} introduced evolutionary computation approaches that are able to produce diverse sets of solutions by evolving a population according to a given quality criteria as well as a diversity measure to the population.

Recently, this approach has been adapted to evolve diverse sets of Traveling Salesperson Problem (TSP) instances~\cite{GaoNN16} as well as diverse sets of images~\cite{DBLP:conf/gecco/AlexanderKN17}. In the case of the TSP, instances have been evolved that are hard to be solved by a given solver. In this case, diversity is measured according to different features that characterize the problem instances. In the case of images, the population of an evolutionary algorithm has been used to evolve images that are close to a given one (in terms of an error measure) and that are diverse with respect to different artistic features. Furthermore, an evolutionary image composition approach based on a feature-based covariance error function has been introduced in~\cite{DBLP:conf/gecco/NeumannSCN17}. Both diversity optimization approaches build on a simple diversity measure that measures diversity according to a given feature. In order to extend this approach to more than one feature, a diversity measure weightening the different features has been used. 

In this paper, we introduce a diversity optimization approach using the discrepancy measure. This approach allows to evolve diverse sets without having any assumption on the preferred weightening of the different diversity criteria.
Discrepancy theory studies the \emph{irregularity of distributions} in the following sense. Given a metric space $S$ and some $n$ points $s_1,\ldots,s_n \in S$, the discrepancy of the set $X:=\{s_1,\ldots,s_n\}$ is measured as the largest deviation from a perfectly evenly distributed point set. When, as in our case, $S=[0,1]^d$ is the $d$-dimensional unit cube, we could measure the discrepancy with respect to all axis-parallel boxes $[a,b]:=[a_1,b_1] \times \ldots \times [a_d,b_d]$. In an ideal situation, we would like the number of points of $X$ that are inside such a box $[a,b]$ to be proportional to its volume. In other words, we would like the difference $\Vol([a,b]) - |X \cap [a,b]|/n$ to be as small as possible, simultaneously for all possible boxes $[a,b]$. The discrepancy is set to be the largest deviation; i.e., 
$$D(X,\mathcal{B}):=\sup\{ \Vol([a,b]) - |X \cap [a,b]|/n \mid a \le b \in [0,1]^d \},$$ 
where we abbreviate $a \le b$ if and only if for every component $i \in d$ the inequality $a_i \le b_i$ holds. The smaller the discrepancy of a point set, the more regular is its distribution with respect to all axis-parallel boxes. 

Discrepancy theory plays an important role in numerical integration, where (under certain circumstances), low discrepancy point sets are known to provide very good estimates for the integral of an unknown or difficult-to-analyze function. Classical Monte Carlo integration is therefore often replaced by a so-called Qusi-Monte Carlo integration, which uses low discrepancy point sets instead of purely random ones, cf.~\cite{Mat99} for an illustrated introduction to discrepancy theory. In the context of evolutionary computation, low discrepancy points sets such as Sobol and Halton sequences have been used in the sampling routines of evolution strategies~\cite{AugerJT05,SchoenauerTT11}, CMA-ES variants~\cite{Teytaud15,Teytaud08,TeytaudG07}, and other genetic algorithms~\cite{KimuraM06,KimuraM05}, and are reported to bring efficiency gains over pure random sampling. On the other hand, evolutionary algorithms have been used to compute point sets of low discrepancy values~\cite{DeRainville2012,DoerrR13}, an optimization problem not admissible by traditional analytical approaches. Finally, randomized search heuristic play also a crucial role for the computation of discrepancy values of point sets in high dimensions~\cite{GWW12}. 

The arguably most intensively studied discrepancy notion is the so-called \emph{star discrepancy}, which measure the regularity with respect to all axis-parallel boxes $[0,b]$, $b \in [0,1]^d$ that are anchored in the origin. This is also the measure for which Sobol and Halton sequences have been designed for. Here in this work, we use this star discrepancy measure to evaluate how evenly the points are distributed. 

At first sight, one might conjecture that a regular $\sqrt{n} \times \sqrt{n}$ grid has a good and regular distribution. Its star discrepancy, however, is rather large: we easily convince ourselves there are boxes of volume $1/\sqrt{n}$ which do not contain any point, so that the star discrepancy is of at least this order. Random point sets also achieve a discrepancy value of order $1/\sqrt{n}$ only. In contrast, the low-discrepancy sequences mentioned above achieve a discrepancy value of order $\log^{d-1}/n$, and are thus much more evenly distributed with in terms of discrepancy.

Apart from numerical integration and the mentioned applications in evolutionary computation, low discrepancy sequences play an important role also in statistics, computer graphics, and stochastic programming.

We investigate the use of the star discrepancy measure in evolutionary diversity optimization for two settings previously studied previously in the literature, namely diversity optimization for images~\cite{DBLP:conf/gecco/AlexanderKN17} and TSP instances~\cite{GaoNN16}. In terms of images, we also introduce a new and more effective mutation operator based on random walks for images than the one introduced in \cite{DBLP:conf/gecco/AlexanderKN17}. This self-adaptive random walk operator allows to reduce the number of iterations needs to construct and good diverse set of solutions from $1-4$ million~\cite{DBLP:conf/gecco/AlexanderKN17} to $2000$ and therefore reduces the number of required generations by $3$ orders of magnitude.

Our experiments are carried out for diversity optimization tasks using two and three features. We show that the previously used approach for images~\cite{DBLP:conf/gecco/AlexanderKN17} and TSP instances~\cite{GaoNN16} computing a weighted diversity contribution in terms of the considered features constructs solution sets with a very high discrepancy compared to our approach using the discrepancy measure. Furthermore, we show that the weighted diversity contribution approach can be used in an effective way for doing tie-breaking between sets of solutions having the same discrepancy value. 

The paper is structured as follows. In Section~\ref{sec:discr}, we introduce our discrepancy-based diversity optimization approach. In Section~\ref{sec:im}, we introduce the new mutation operator for diversity optimization of images and evaluate the discrepancy optimization approach for images this approach for images.
We consider our approach for evolving sets of TSP instances of low discrepancy with respect to the given features in Section~\ref{sec:tsp}. Finally, we finish with some concluding remarks.

\section{Discrepancy-based Diversity Optimization}
\label{sec:discr}

\begin{algorithm}[t]
{
   	Initialize the population $P$ with $\mu$ instances of quality at least $\alpha$.\\
	Let $C \subseteq P$ where $|C| = \lambda$.\\
	For each $ I \in C$, produce an offspring $I'$ of $I$ by mutation. If $q(I')\geqslant\alpha$, add $I'$ to $P$. \\
	While $|P| > \mu$, remove an individual $I=\arg \min_{J \in P} D^*(P\setminus J)$.\\

 	Repeat step 2 to 4 until termination criterion is reached.\\
} 
 \caption{$(\mu+\lambda)$-$EA_{D}$}
\label{EA}
\end{algorithm}
We consider evolutionary diversity optimization.
Given a search space $S$, our aim is to construct a diverse set of solutions $P=\{X_1, \ldots, X_{\mu}\}$ where each solution $X_i \in S$ fullfills a given quality criteria, i.e. we have $q(X_i) \geq \alpha$ for a given quality threshold $\alpha$. 

Properties of our potential solutions $X_i$ are characterized by features $f_1, \ldots, f_d$ which are problem specific.
Let $I \in S$ be an individual in a population $P$. With associate with $I$ its feature vector $f(I) = (f_1(I), \ldots, f_d(I))$. 

Traditionally, the goal of construction a set of points with a low discrepancy is defined in $[0,1]^d$.
Therefore, the feature values are scaled before the calculation of discrepancy. Let $f_i^{\max}$ and $f_i^{\min}$ be the maximum and minimum value of feature $f_i$. 
We evaluate our set of points in terms of discrepancy using the scaled feature values
$$f_i'(I) = (f_i(I)-f_i^{\min})/(f_i^{\max}-f_i^{\min}).$$ 
We have $f'(I) \in [0,1]^d$ for all scaled feature vectors $f'(I)$ iff $f_i^{\min} \leq f_i(I) \leq f_i^{\max}$, $1 \leq i \leq d$. $f_{\max}$ and $f_{\min}$ are set based on initial experiments. Feature values outside that range would be scaled to $0$ and $1$, respectively, to allow the algorithm to work with non anticipated features values.

Let $f'(P) = \cup_{I \in P} f'(I)$ be the (multi-)set of (scaled) features vector in $P$
We denote by $D^*(P)$ the discrepancy of $f'(P)$ in $[0,1]^d$. Throughout this paper, we use the star-discrepancy. Given $P=\{I_1, \ldots, I_k)$ with feature vectors $f'(I_1), \ldots, f'(I_k)$, we define
\[
D^*(P) = sup_{J \in Y} D(J, P)
\]
with 
\[
D(J,P) = \frac{|I \in P \mid f'(I) \in J |}{k} - Vol(J).
\]
Here Vol(J) denotes the volume of interval $J$ and
$Y$ is the class of all subintervals of the form
\[
J = \prod_{i=1}^d [0, u_i)
\]

with $0 \leq u_i \leq 1$ for $1 \leq i \leq d$.

\begin{figure}
\centering
\includegraphics[width=0.28\textwidth]{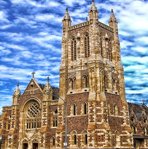}
\caption{Image S.}
\label{fig:ImageS}
\end{figure}

A key difficulty to overcome in the optimization for low star discrepancy values is the computational hardness of its evaluation~\cite{GnewuchSW09}. The best known algorithm for the star discrepancy computation has a running time of order $n^{1+d/2}$~\cite{DobkinEM96}, which is exponential in the dimension $d$. As we are interested in dimension $d=2, 3$, we can use this algorithm, and make use of the  implementation that is available on~\cite{Wahl12}. The reader interested in a discussion of computational aspects of geometric discrepancies, along with a description of the above-mentioned heuristic approach for its approximation, is referred to~\cite{DoerrGW12}.

We use the $(\mu+\lambda)$-$EA_{D}$ given in Algorithm~\ref{EA} to compute a diverse population where each individual meets a given quality criteria $q$ according to a given threshold $\alpha$, i.e. we have $q(I) \geq \alpha$ for all individuals in the population $P$.
The population $P$ is a multi-set, i.e. it may contain an instance more than once.
The algorithm is initialized with a population where each individual meets the given criteria. In each iteration $\lambda$ offspring are produced. Offspring that do not meet the quality criteria are directly rejected. Offspring that meet the criteria are added to the population and survival selection is performed afterwards to obtain a population of size $\mu$. To do this, individuals are removed iteratively. Having a population of size $k>\mu$, in each iteration an individual $I$ is removed that leads to a population $P \setminus I$ of size $k-1$ having the smallest discrepancy among all populations that can be constructed by removing exactly one individual from $P$.

The discrepancy minimization algorithm is compared to evolutionary diversity optimization approach in ~\cite{GaoNN16} which aims maximizing the feature-based population diversity using a weighted contribution measure for each individual. The weighted diversity contribution of an individual $I$ with feature vector $f(I)$ is defined as
$$c(I,P)=\sum^{k}_{i=1}(w_i\cdot d_{f_i}(I,P)),$$

where $d_{f_i}(I,P)$ represents the normalised contribution of individual $I$ to the population diversity over feature $f_i$ and $w_i$ represents the weight for feature $f_i$.

The resulting algorithm $(\mu+\lambda)$-EA$_{C}$ differs from $(\mu+\lambda)$-EA$_{D}$ only in step 4), and removes in each of these steps an individual $I$ with the smallest weighting contribution $c(I,P)$ to the population diversity. 
Furthermore, we consider the algorithm $(\mu+\lambda)$-EA$_{T}$ which uses both the discrepancy measure the weighted contribution measure. It is the same as $(\mu+\lambda)$-EA$_{D}$ but uses the weighted contribution measure as tie-breaking in step 4) of the algorithm, i.e. if there is more than one individual whose removal leads to the minimum discrepancy value than the one among them with the smallest contribution to weighted contribution diversity measure is removed. 

In the following, we evaluate our discrepancy-based diversity optimization approaches for evolving diverse sets of images and TSP instances. We also introduce a new mutation operator for images based on random walks which significantly speeds up the diversity optimization process when constructing a diverse set of images.

\section{Images}
\label{sec:im}

\begin{figure}[t]
\centering
\includegraphics[width=0.24\textwidth]{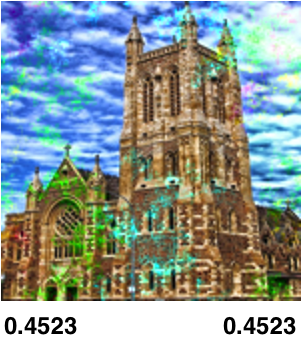}
\includegraphics[width=0.24\textwidth]{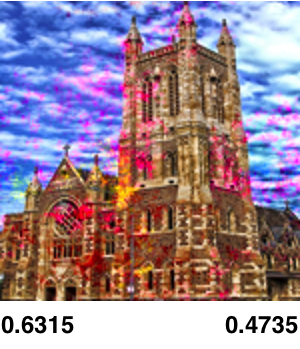} 
\includegraphics[width=0.24\textwidth]{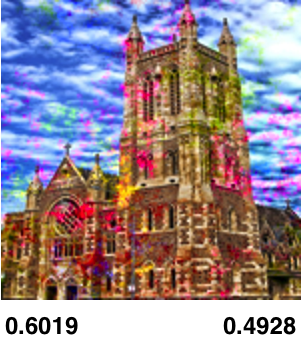}
\includegraphics[width=0.2418\textwidth]{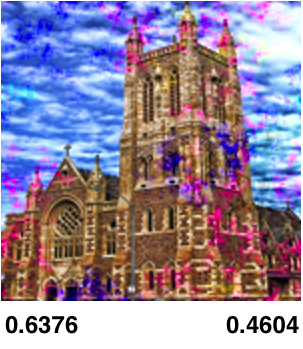}
\caption{Selected images from the population after discrepancy minimization for the Hue and Saturation features.}
\label{hue:saturation}
\end{figure}
\begin{figure*}
\centering

\includegraphics[width=0.32\textwidth]{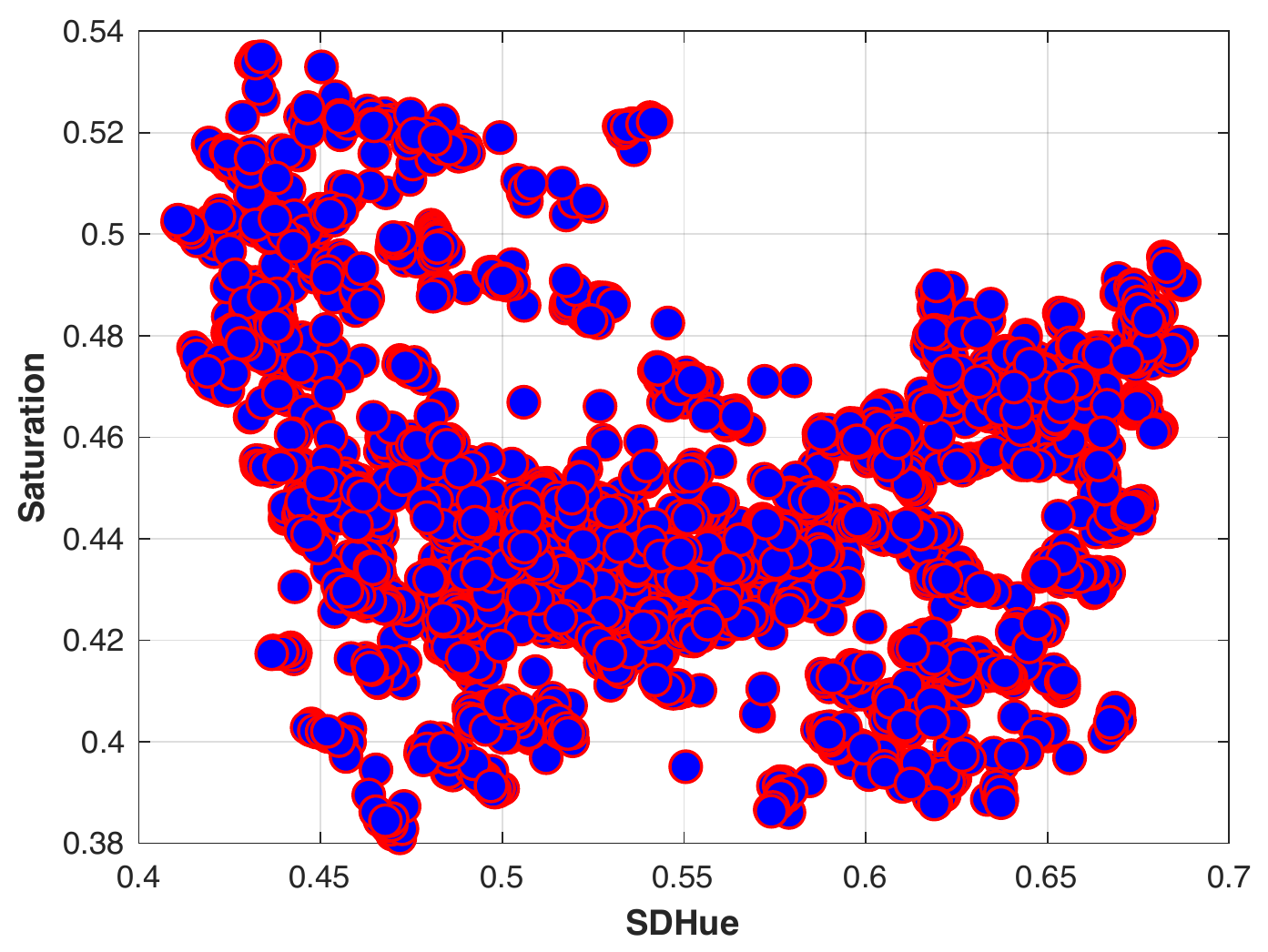} 
\includegraphics[width=0.32\textwidth]{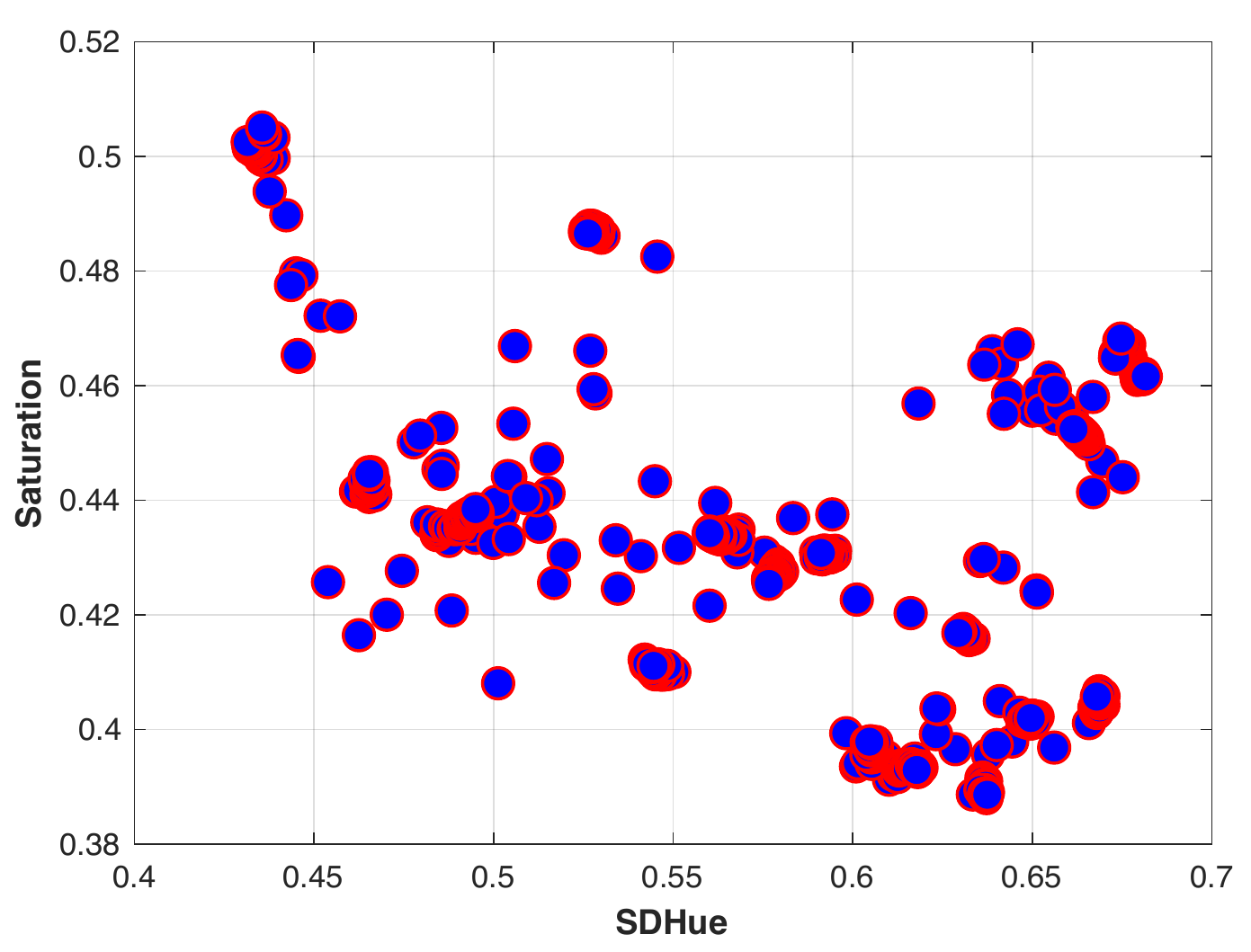}
\includegraphics[width=0.32\textwidth]{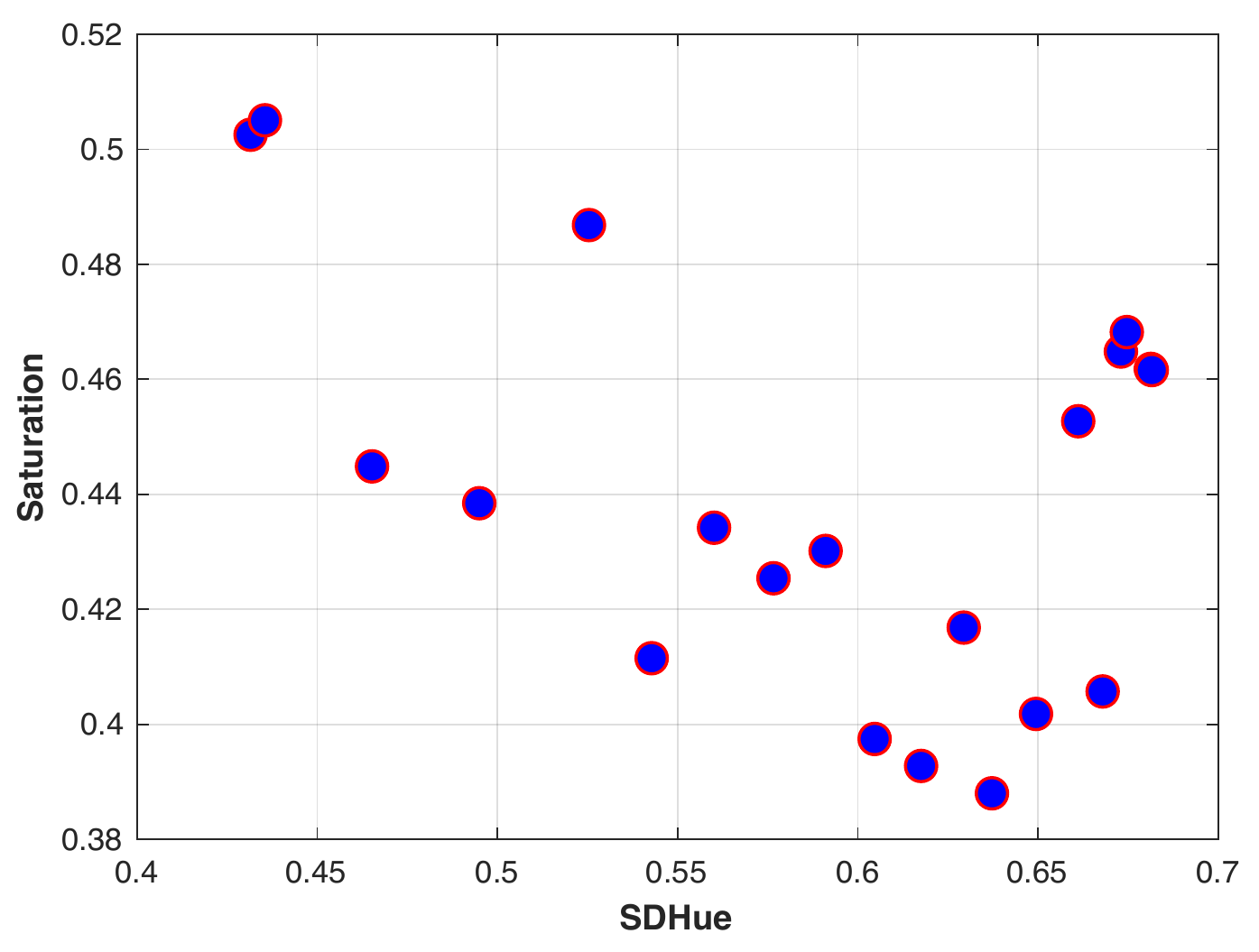}

\caption{All feature vectors generated in 10 runs of $(\mu+\lambda)$-$EA_{T}$ with 1000 iterations each (left), one run with 1000 iterations (middle), the final population after 1000 iteration with discrepancy 0.22637 (right).}
\label{plot:all_runs}
\end{figure*}

We consider the task of evolving a diverse set of images as previously investigated in~\cite{DBLP:conf/gecco/AlexanderKN17}.
Given an image $S$, the task is to compute a diverse set of images $P=\{I_1, \ldots, I_{\mu}\}$ that meets a given quality criterion $q(I)$ for each $I \in P$.
For our experimental investigations, an image $I$ meets the quality criterion if the \emph{mean-squared error} in terms of the RGB-value of the pixels of the image $I$ with respect to the input image $S$ (shown in Figure~\ref{fig:ImageS}) is less than $500$.

Many features have been used to measure the characteristics of images. We focus on a selected set of features used in~\cite{DBLP:conf/gecco/AlexanderKN17,den2014investigating}.
We carry out our discrepancy-based diversity optimization approach for different features and utilised the evolutionary algorithm to evolve  diverse populations of images for each feature combination.
The set of features used in our experiments are as follows: 
{\em{standard-deviation-hue}}, {\em{mean-saturation}}, {\em{reflectional symmetry}}~\cite{den2014investigating}, {\emph{mean-hue}}, {\emph{Global Contrast Factor (GCF)}}~\cite{matkovic2005global} and 
{\em{smoothness}}~\cite{Nixon:2008:FEI:1571711}.

We focus our experiments on the characterization of how the chosen features may influence the generated images. In reference to previous work~\cite{DBLP:conf/evoW/NeumannAN17} we choose three pairs of features: {\em{standard-deviation-hue}} and {\em{mean-saturation}}, {\em{symmetry}} and {\emph{mean-hue}}, {\emph{GCF}} and {\em{smoothness}}. 
In addition, we choose three feature sets consisting of three features each as follows: {\em{standard-deviation-hue}}, {\em{mean-saturation}}, {\em{symmetry}}, and {\em{standard-deviation-hue}}, {\emph{mean-hue}}, {\em{symmetry}}, and {\emph{GCF}}, {\emph{mean-hue}}, {\em{mean-saturation}}.

We are working with the scaled feature values when computing the discrepancy of a given set of points. It should be pointed out that not all feature vector combinations within the given feature intervals are usually possible. To illustrate this we consider the features SDHue and Saturation and run the EA (using the mutation operator described in Section~\ref{sec:mut}) for $1000$ iterations. 
Figure~\ref{plot:all_runs} shows all feature vectors produced during 10 runs of the $(20+1)$-$EA_{T}$ (left), all feature vectors produced during one run (middle), and the feature vectors of the final population (right). It can be observed that the area where both feature values are high does not contain any points (similarly if both feature values are very low). The seems to indicate that the problem is constrained to a subspace of the unit square. If this is true, then this has a direct consequence on the best possible discrepancy value that can be obtained, as discrepancy is a measure in $[0,1]^d$.

\subsection{Self Adaptive Offset Random Walk Mutation}
\label{sec:mut}

\begin{algorithm}[t]
    Let $X$ is a image with pixels $X_{ij} \in X$.
    
    $Y \gets X$.
    
Choose starting pixel  $Y_{ij} \in  Y$ uniformly at random.
    
Choose offset $o \in [-r, r]^3$ uniformly at random.

    $t \gets 1$.
    
    \While {$t \le \tmax$}{
    
   $Y_{ij} = Y_{ij} + o$.
    
    Choose $Y_{kl} \in N(Y_{ij})$ uniformly at random.
    
   $i \gets k$, $j \gets l$.
    
    $t=t+1$.
    }
     Return $Y$.
\caption{\textsc{OffsRandomWalkMutation ($X,\tmax$)}}
\label{alg:randomwalkcolour}
\end{algorithm}

The algorithm uses a variant of the random walk mutation introduced in~\cite{DBLP:conf/gecco/NeumannSCN17} for evolutionary image composition.
This speeds up the process of diversity optimization by three orders of magnitude compared to \cite{DBLP:conf/gecco/AlexanderKN17} where for a mutation operator changing in each step a single pixel $1-4$ million iterations where required to construct a diverse set of images. Our new mutation operator enables us to construct diverse sets of images for all three algorithms (including $(\mu+\lambda)$-EA$_{C}$ investigated in \cite{DBLP:conf/gecco/AlexanderKN17}) within just $2000$ generations.

The random walk in this paper differs from the one for image composition given in \cite{DBLP:conf/gecco/NeumannSCN17} by changing the $RGB$ values by an offset vector $o \in [-r,r]^3$ chosen in each mutation step uniformly at random. 
The mutation operator is shown in the Algorithm~\ref{alg:randomwalkcolour}.

The random walk causes movement from the current pixel $X_{ij}$ to the next pixel by moving either right, left, down or up. 
We define the 
neighbourhood $N(X_{ij})$ of pixel $X_{ij}$ as \begin{displaymath}
  N(X_{ij}) = \left\{ X_{(i-1)j}, X_{(i+1)j}, X_{i(j-1)}, X_{i(j+1)} \right\}.
\end{displaymath}

\begin{figure*}
\centering
\rotatebox{90}{\hspace{6mm}$(\mu+\lambda)$-EA$_{C}$} \rotatebox{90}{\rule{33mm}{1pt}}
\includegraphics[width=0.3\textwidth]
{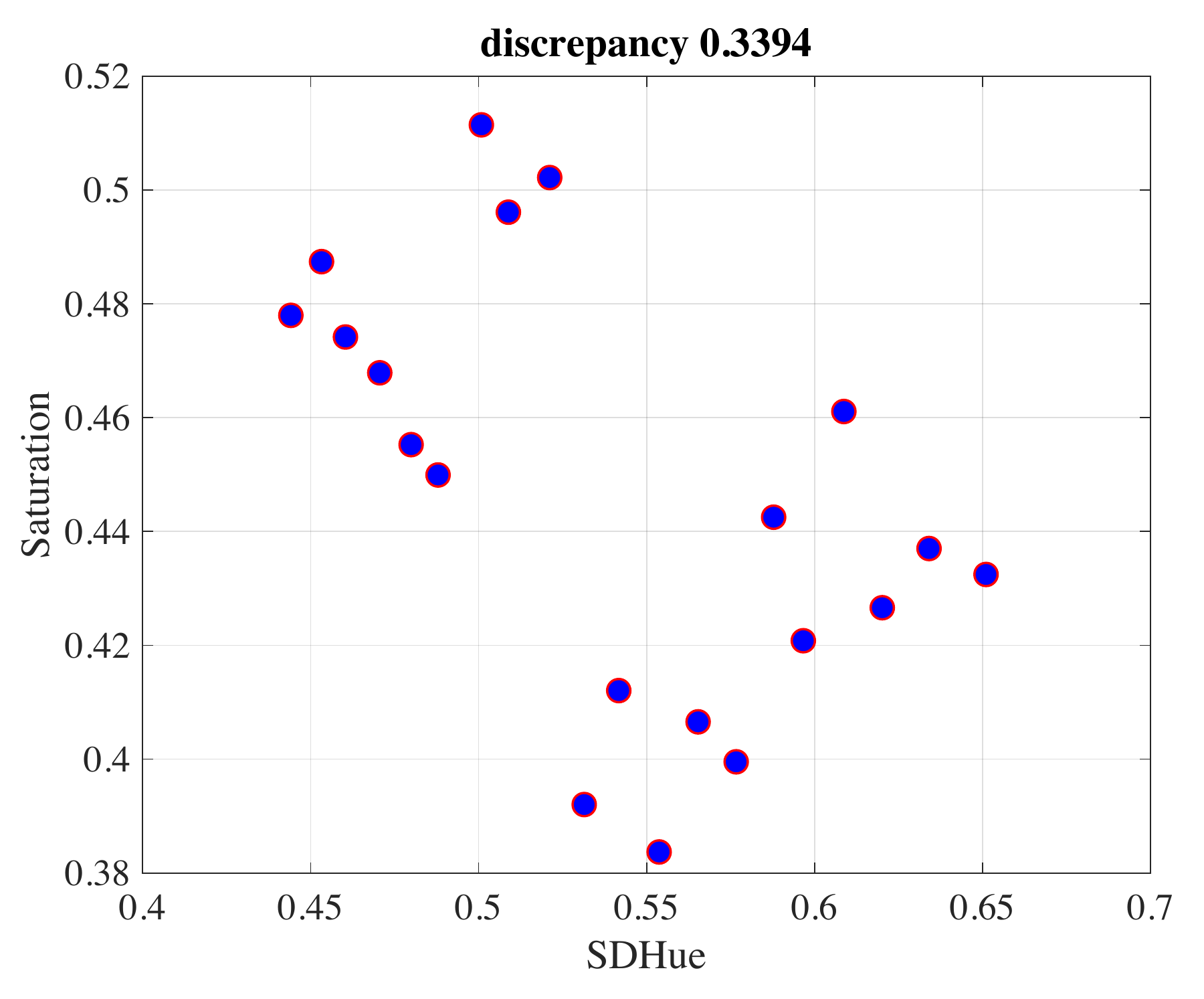}
\includegraphics[width=0.3\textwidth]
{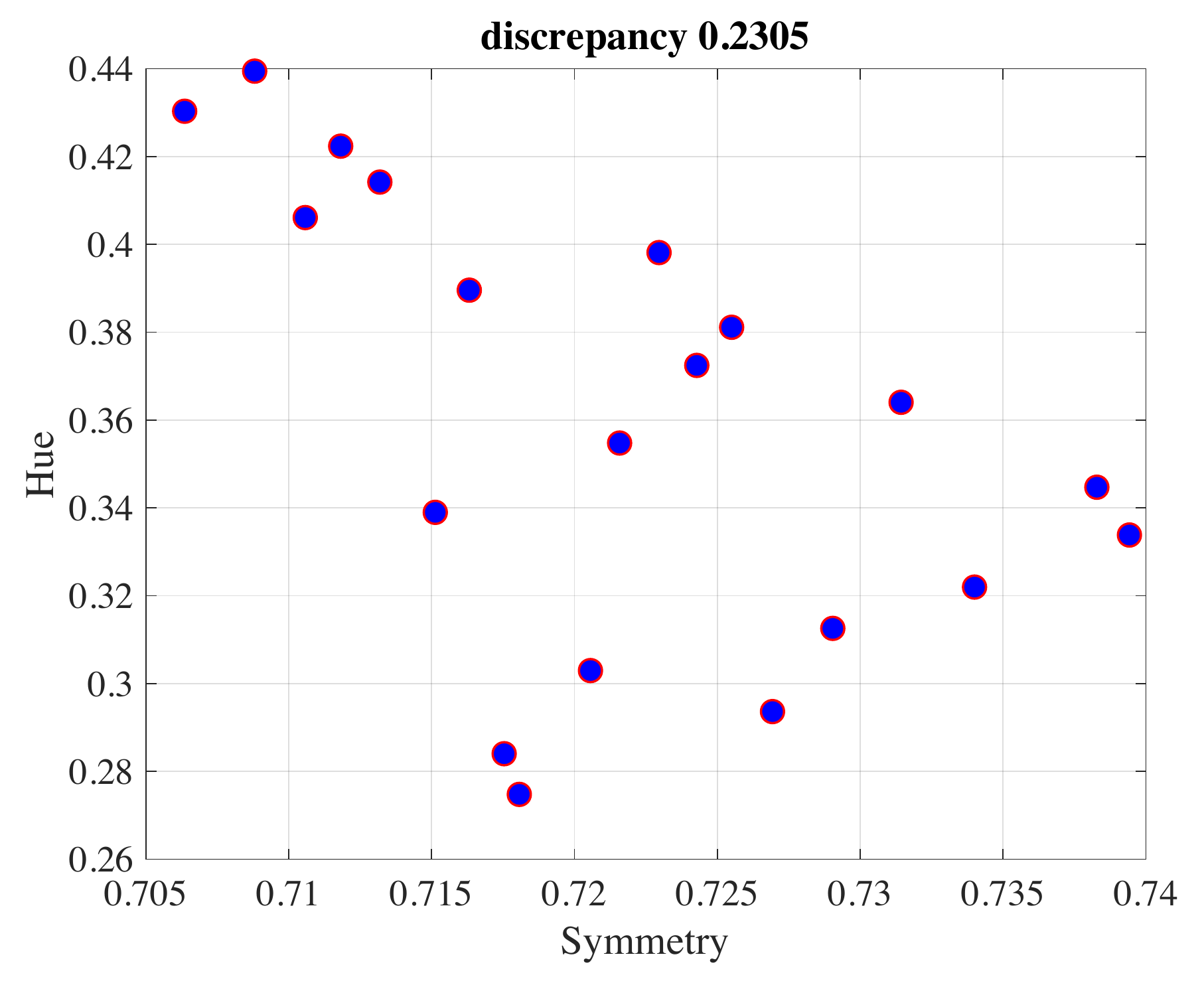}
\includegraphics[width=0.313\textwidth]
{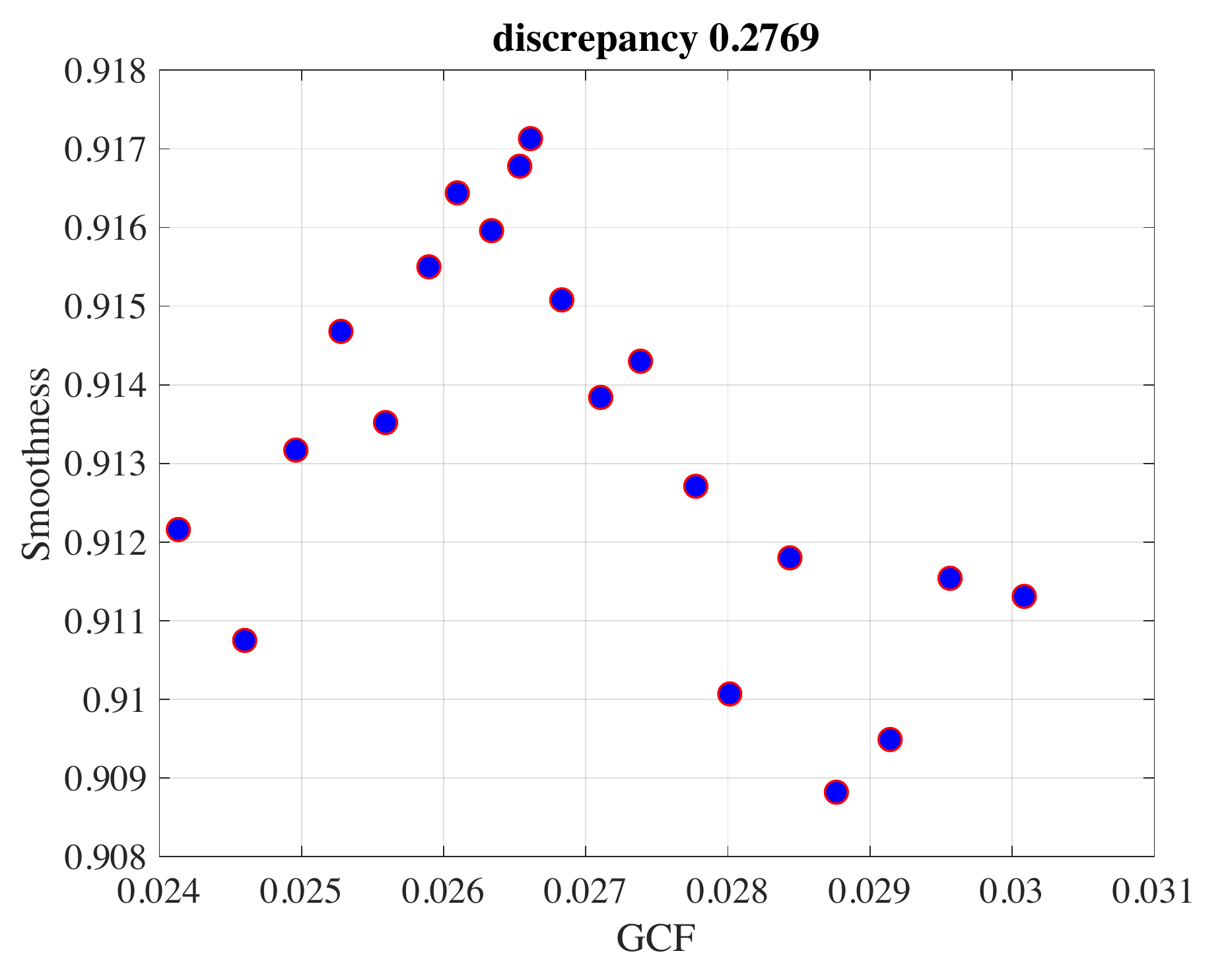}\\

{ \fontsize{3.5}{5}\selectfont \bf  
\hspace{43mm} 
{\fontfamily{ptm}\selectfont discrepancy 0.1544 }\hspace{28.5mm}   {\fontfamily{ptm}\selectfont discrepancy 0.1366 }}\\ \vspace{-0.755cm}
\vspace{3mm}\rotatebox{90}{\hspace{6mm}$(\mu+\lambda)$-EA$_{D}$} \rotatebox{90}{\rule{33mm}{1pt}}
\includegraphics[width=0.3\textwidth]
{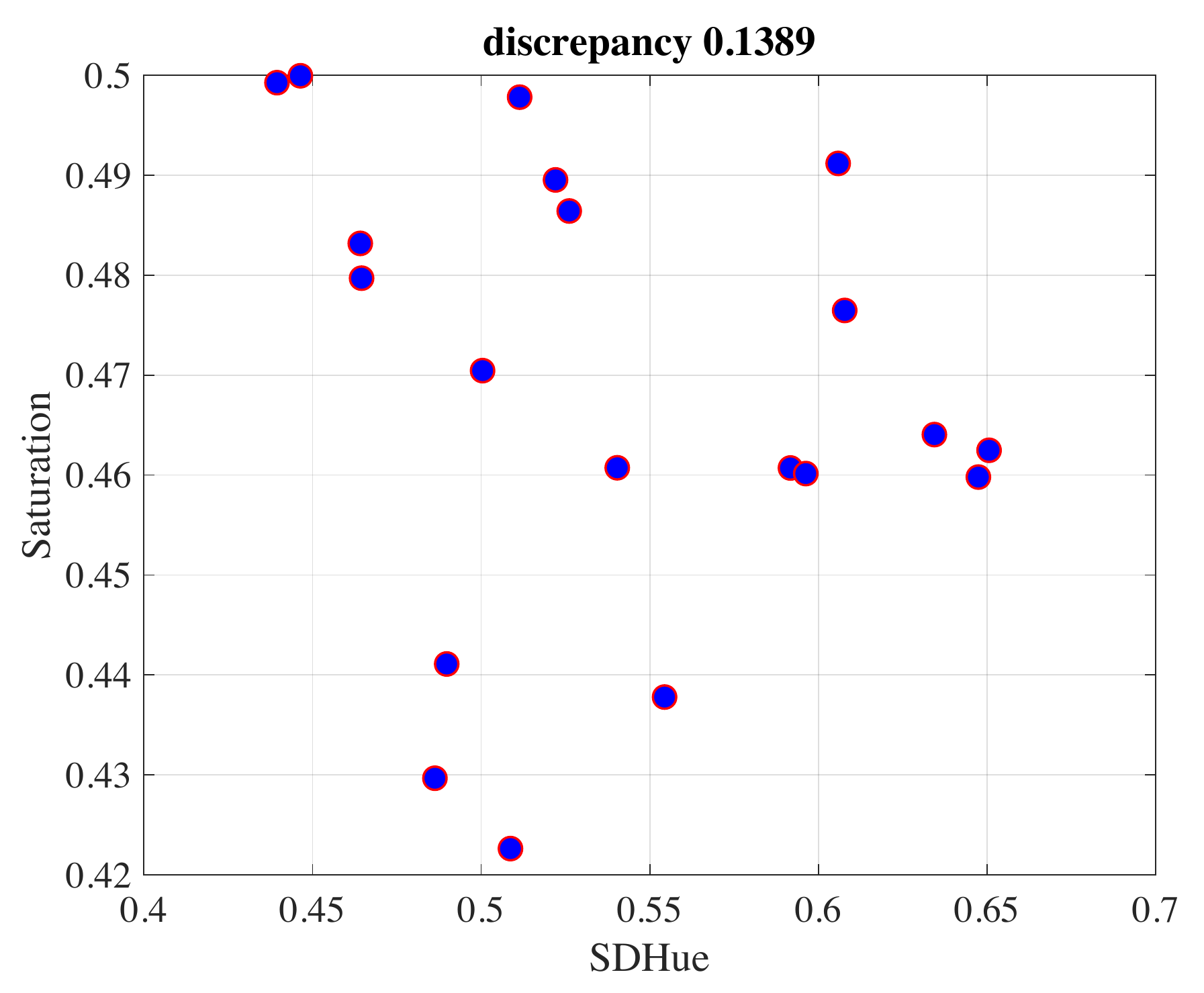}
\includegraphics[width=0.3\textwidth, height=0.15\textheight]{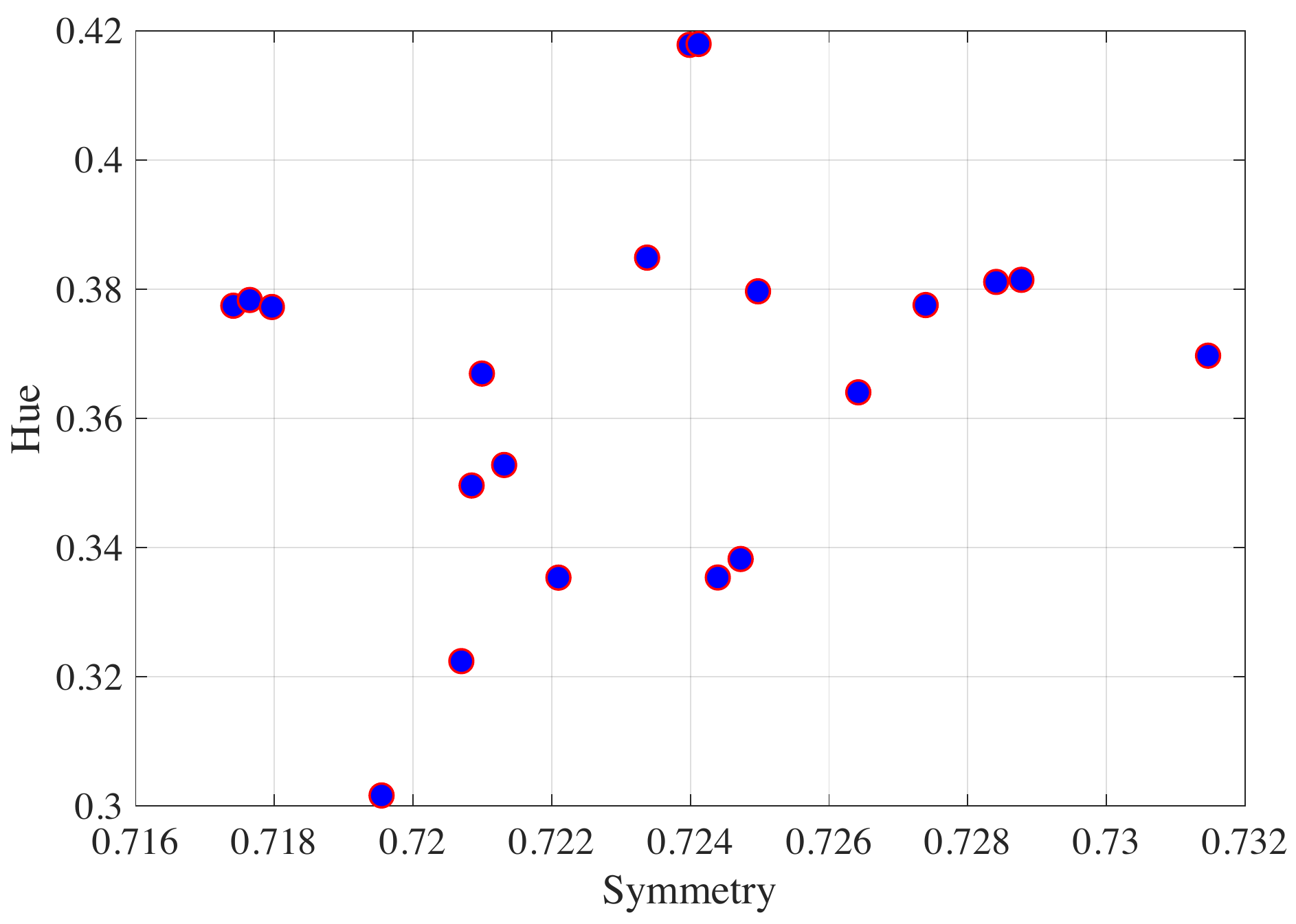}
\includegraphics[width=0.309\textwidth, height=0.15\textheight]{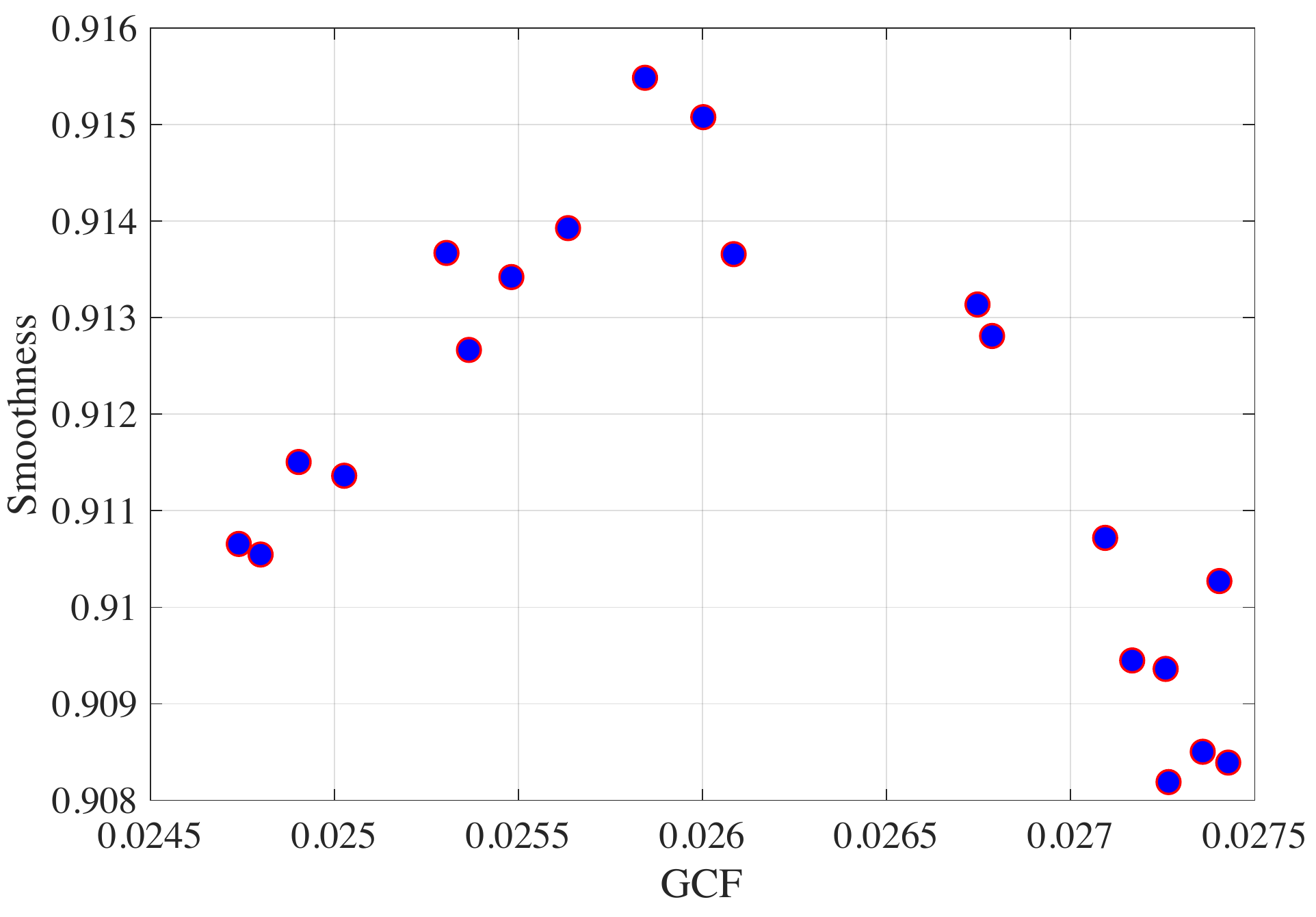}\hspace*{0.6mm}

\caption{Feature vectors for final population of $(\mu+\lambda)$-EA$_{C}$ (top) and  $(\mu+\lambda)$-EA$_{D}$ (bottom) for Images based on two feature from left to right: (SDHue, Saturation), (Symmetry, Hue), (GCF, Smoothness).
} 
\label{plot:fig1}
\end{figure*}

The random walk chooses an element of $N(X_{ij})$ uniformly at random in every step. Furthermore, the random walk is wrapped around the boundaries of the image. We produce an offspring $Y$ from $X$ by setting each visited pixel $X_{ij}$ to the value of $X_{ij} + o$.
Given a current image $X$, our $\mlea$ algorithm uses the random walk mutation to alter all visited pixels. Note that pixels may be visited more than once and the offset may be applied several times in this case. 
The random walk paints all the visited pixels by adding the chosen offset vector $o$. Each random walk mutation is run for \tmax steps, where \tmax is chosen in an adaptive way.

\subsubsection{Self Adaptation} 

We decrease the length of random walks through decreasing \tmax when the discrepancy value does not decrease as a result of an unsuccessful mutation. We increase \tmax if the discrepancy decreases as a result of a successful mutation. This builds on the assumption that mutations doing less change to the image are needed to obtain an improvement if it is hard to make progress with the current choice of \tmax. On the other hand, a larger progress may be achievable if the current setting of \tmax is already able to decrease the discrepancy. 
Our adaptive approach makes use of the parameter adjusting scheme that was recently introduced by Doerr and Doerr
~\cite{DBLP:conf/gecco/DoerrD15}.
Their method modifies the classical $1/5$-rule adaptation for evolution strategies in \cite{DBLP:conf/gecco/Auger09} to a discrete setting. 

Our approach increases \tmax for a successful outcome or decreases \tmax in the case that the new offspring is not accepted. In our algorithm, \tmax can take on values in
 ${t_{\mathrm{LB}} \leq t_{\max} \leq t_{\mathrm{UB}}}$, where $t_{\mathrm{LB}}$ is a lower bound on \tmax and $t_{\mathrm{UB}}$ is an upper bound on \tmax.

For a successful mutation, we set
\begin{displaymath}
  t_{\max} \colonequals \min \left\{F\cdot t_{\max}, \, t_{\mathrm{UB}} \right \} 
\end{displaymath}
and for an unsuccessful mutation, we set
\begin{displaymath}
  t_{\max} \colonequals \max \left\{F^{-1/k} \cdot t_{\max}, \, t_{\mathrm{LB}} \right \},
\end{displaymath}
where $F>1$ is a real value and $k\geq 1$ an integer which determines the adaptation scheme.

For our experimental investigations, we set $t_{\mathrm{LB}}=1000$, $t_{\mathrm{UB}}=20000$, $F=2$, $k=8$, and $\tmax=1000$ at initialization based on preliminary experimental investigations.

\subsection{Experimental settings}

All algorithms were implemented in $Matlab$ $(R2017b)$. We ran all of our experiments on single nodes of a Lenovo NeXtScale M5 Cluster with two Intel Xeon E5$-$2600 v4 series 16 core processors, each with 64GB of RAM.

Firstly, we consider the discrepancy-based diversity optimization for two features.
We select features in order to combine different aesthetic and general features based on our initial experimental investigations and previous investigations in~\cite{DBLP:conf/evoW/NeumannAN17}. Furthermore, we set for each feature value in the specifize range.
Note for $SDHue$, $Hue$, $Saturation$, $Smoothness$, $GCF$, $Symmetry$ the $f^{min}$ values are $0.42$, $0.25$, $0.42$ , $0.42$, $0.906$, $0.0245$, and $0.715$, respectively. The corresponding $f^{max}$ values are $0.7$, $0.4$, $0.5$, $0.5$, $0.918$, $0.0275$, and $0.74$.

After having consider the combination of two features, we investigate sets of three features.
Here, we selected different features combining aesthetic and general features together used in the previous experiment. In order to obtain a clear comparison between experiments, we applied the same range of feature values as before.

Furthermore, we run the $(\mu+\lambda)$-$EA_{C}$ diversity algorithm from~\cite{DBLP:conf/evoW/NeumannAN17} using self adaptive random walk mutation operator in order to compare the two approaches for diversity optimization. We utilized the same settings for the $(\mu+\lambda)$-$EA_{C}$ as we done for our discrepancy-based diversity algorithm $(\mu+\lambda)$-$EA_{D}$. Finally, we consider $(\mu+\lambda)$-$EA_{T}$ which uses discrepancy-based diversity optimization plus tie-breaking according to weighted feature contributions when more than one individual would lead to the minimal discrepancy value obtainable by removing exactly one individual.

We run each algorithm for $2000$ generations with a population size of $\mu = 20$ and $\lambda=1$.
In order to evaluate our results using statistical studies, each algorithm has been run 30 times with the same setting applied to each considered pair and triple of features.

\subsection{Experimental Results}

We perform a series of experiments to evaluate the performance of our discrepancy-based diversity evolutionary algorithm.
Our experiments establish that global constraints like \emph{mean-squared error} can be used to produce more diverse images than equivalent constraints which are limited to the range of the colour or luminosity channel for each pixel.

The images displayed in Figure~\ref{hue:saturation} match the color features. In particular, images
are produced using the $\sdhue$ and $\sat$ feature. The values for $\sdhue$ and $\sat$ are displayed above each image for features, respectively.
The color spectrum is red at each end with individuals of the population spread along it.
Images which have a low score for this feature will be monochromatic and will appear in the middle of the spectrum.
Images with a high score will be read as it is a sample from both of the extremes.
Low-scoring images in the $\sat$ feature are monochromatic whilst high-scoring individuals are almost entirely saturated.

Figure~\ref{plot:fig1} shows feature plots of the final populations of $(\mu+\lambda)$-$EA_{C}$ (top) and $(\mu+\lambda)$-$EA_{D}$ (bottom) for 3 pairs of feature combinations.
It can be observed that the the discrepancy value for $(\mu+\lambda)$-$EA_{D}$ and the combination $(SDHue,\sat)$ is $0.1389$. This is significantly smaller than the one for $(\mu+\lambda)$-$EA_{C}$ at $0.3394$.
The middle row shows the combination of$\hue$ and $\symm$. The discrepancy value of $\symm$ and $\hue$ for $(\mu+\lambda)$-$EA_{D}$ is $0.1544$ whereas it is $0.2305$ for $(\mu+\lambda)$-$EA_{C}$.
In Figure~\ref{plot:fig1} the right columns shows the final populations of the diversity optimization when considering $GCF$ and $\smooth$. The discrepancy value for $\GCF$ and $\smooth$ is $0.1366$ for $(\mu+\lambda)$-$EA_{D}$ and  $0.2769$ for $(\mu+\lambda)$-$EA_{C}$.
This an indication of the difficulty of evolving images which are smooth as well as scoring high in $\GCF$ in our current setup.
However, this conflict is expected as the $\GCF$ highly scores in the case of strong contrast between adjacent pixels. The $\smooth$ scores have a high value for low contrast between neighbouring pixels.

We now consider the results of $(\mu+\lambda)$-$EA_{C}$  from~\cite{DBLP:conf/evoW/NeumannAN17} using self adaptive random walk mutation operator in greater detail. Looking at Figure~\ref{plot:fig1} (top) which shows the population of instances for $\sdhue$ and $\sat$, $\symm$ and $\hue$, and $\GCF$ and $\smooth$, respectively, we can observe that the distribution of the points for the features vectors for final population follows a linear pattern. This is due to the chosen weights which favor lines of feature vector orthogonal to the used weight vector $(1,1)$.

In the Table~\ref{tb:images} we provide statistics on the discrepancy values for the final populations of $(\mu+\lambda)$-$EA_{C}$, $(\mu+\lambda)$-$EA_{D}$ and $(\mu+\lambda)$-$EA_{T}$, respectively. 
For each algorithm and feature combination the minimum, mean, and standard deviation of the discrepancy value of the final population of $30$ runs is shown.
$(\mu+\lambda)$-$EA_{D}$ clearly outperforms the $(\mu+\lambda)$-$EA_{C}$ for all feature combinations. Furthermore, $(\mu+\lambda)$-$EA_{T}$ which uses tie-breaking according to weighted feature contribution leads to a further improvement of $(\mu+\lambda)$-$EA_{D}$ for almost all feature combinations in terms of the mean and minimal discrepancy value achieved within $30$ runs

\section{TSP}
\label{sec:tsp}
Another problem we considered as application of Algorithm~\ref{EA} is the Travelling Salesman Problem (TSP), which is a NP-hard combinatorial optimization problem with many real world applications. We consider the classical Euclidean TSP which takes a set of cities in the Euclidean plane and the goal is to find a Hamiltonian cycle with the minimal sum of distance.

In this research we focus on TSP instances with 50 cities in the space of $[0,1]^2$ which is a reasonable size of problem for feature analysis of TSP. The instances are qualified with respect to the approximation ratio, which is calculated by  
\[
  \alpha_A(I) = A(I) / OPT(I)
\]
where $A(I)$ is value of the solution found by algorithm $A$ for the given instance $I$, and $OPT(I)$ is value of an optimal solution for instance $I$ that is calculated using the exact TSP solver Concorde~\cite{Applegate02}. Within this study, $A(I)$ is the tour length obtained by three independent repeated runs of the 2-OPT algorithm for a given TSP instance $I$.

Following the same setting as in~\cite{GaoNN16}, the approximation ratio threshold for hard TSP instance of size 50 is set to 1.18, which means only instances with approximation ratio equal or greater than 1.18 are accepted into the population.

The analysis of TSP in feature space attracts more and more attention. There has been many studies on the relationship between problem hardness and the feature values for TSP~\cite{SmithMiles2010,Mersmann2013}.

\subsection{Experiments settings}

There have been many features designed for the TSP with the aim of describing the hardness and characteristics of a certain TSP instance. In this paper, the focus is on a selected set of feature values from the paper of ~\cite{Mersmann2013}. 

According to the experimental results from the previous research into the relationship of TSP feature and problem hardness by ~\cite{GaoNN16}, we focus on combinations of the following features of TSP which gives better indication about problem hardness:

\begin{itemize}
\item \emph{angle\_mean}: the mean value of the angles made by each point with its two nearest neighbor points
\item \emph{mst\_depth\_mean}: the mean depth of the minimum spanning tree in the TSP
\item  \emph{centroid\_mean\_distance\_to\_centroid}: the mean value of the distances from the points to the centroid
\item \emph{mst\_dists\_mean}: the mean distance of the minimum spanning tree
\end{itemize}

As mentioned in Section~\ref{sec:discr}, the feature values are normalized before discrepancy calculation. The maximum and minimum values $f^{max}$ and $f^{min}$ for each feature are determined based on the results gathered from initial runs of feature-based diversity maximization. The $f^{max}$ used for the feature angle\_mean, centroid\_mean\_dist\_centroid, nnds\_mean and mst\_dists\_mean are 2.8, 0.6, 0.7 and 0.15, respectively. The corresponding $f^{min}$ values are 0.8, 0.24, 0.1 and 0.06.
\begin{figure*}
\label{fig:TSP2D}
\rotatebox{90}{\hspace{4.5mm}$(\mu+\lambda)$-EA$_{C}$} \rotatebox{90}{\rule{28mm}{1pt}}
\includegraphics[trim={0.5cm 7.7cm 0.7cm 6.8cm},clip,width=0.3\textwidth]{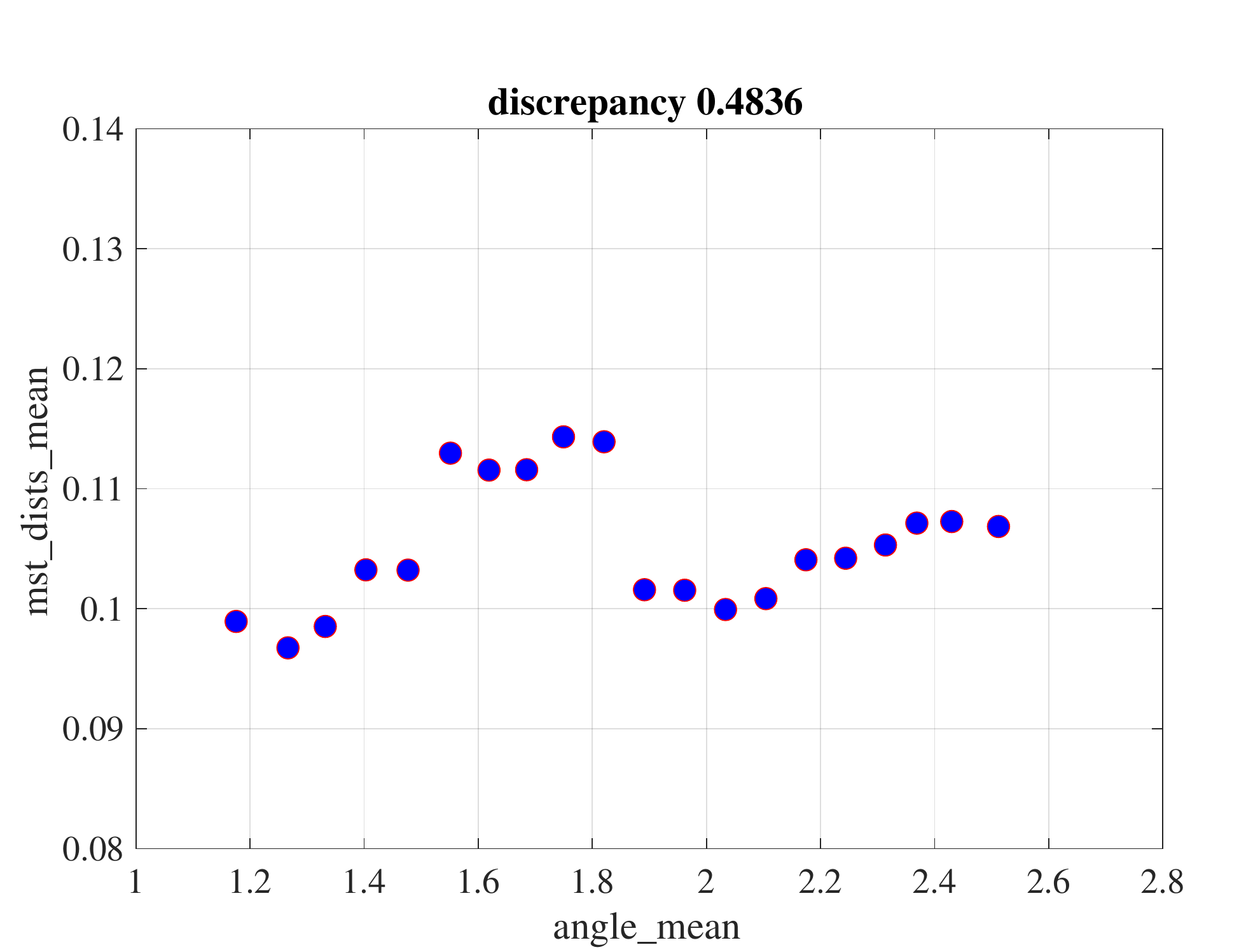}
\includegraphics[trim={0.5cm 7.7cm 0.7cm 6.8cm},clip,width=0.3\textwidth]{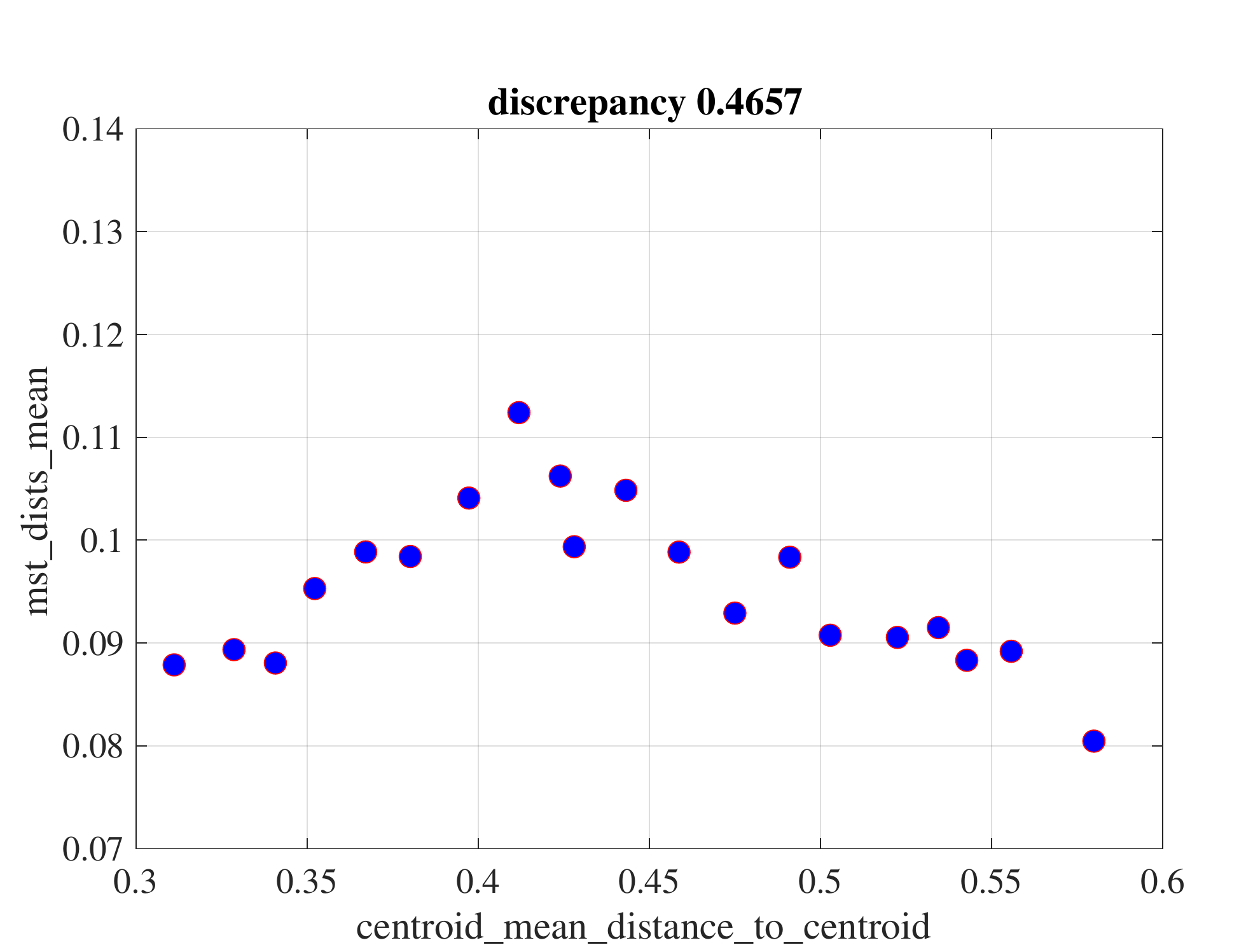}
\includegraphics[trim={0.5cm 7.7cm 0.7cm 6.8cm},clip,width=0.3\textwidth]{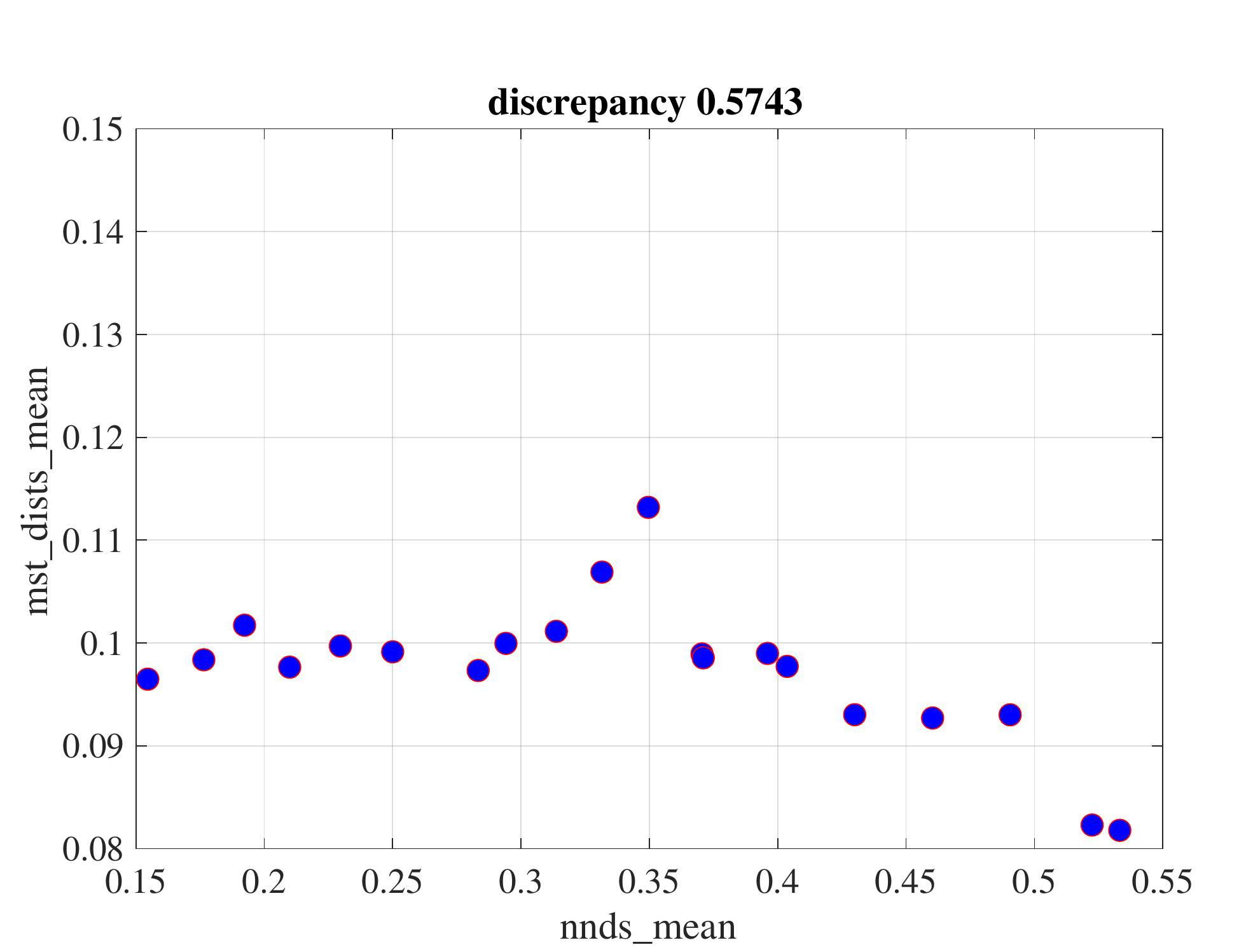}\\
\rotatebox{90}{\hspace{4.5mm}$(\mu+\lambda)$-EA$_{D}$} \rotatebox{90}{\rule{28mm}{1pt}}
\includegraphics[trim={0.5cm 7.7cm 0.7cm 6.8cm},clip,width=0.3\textwidth]{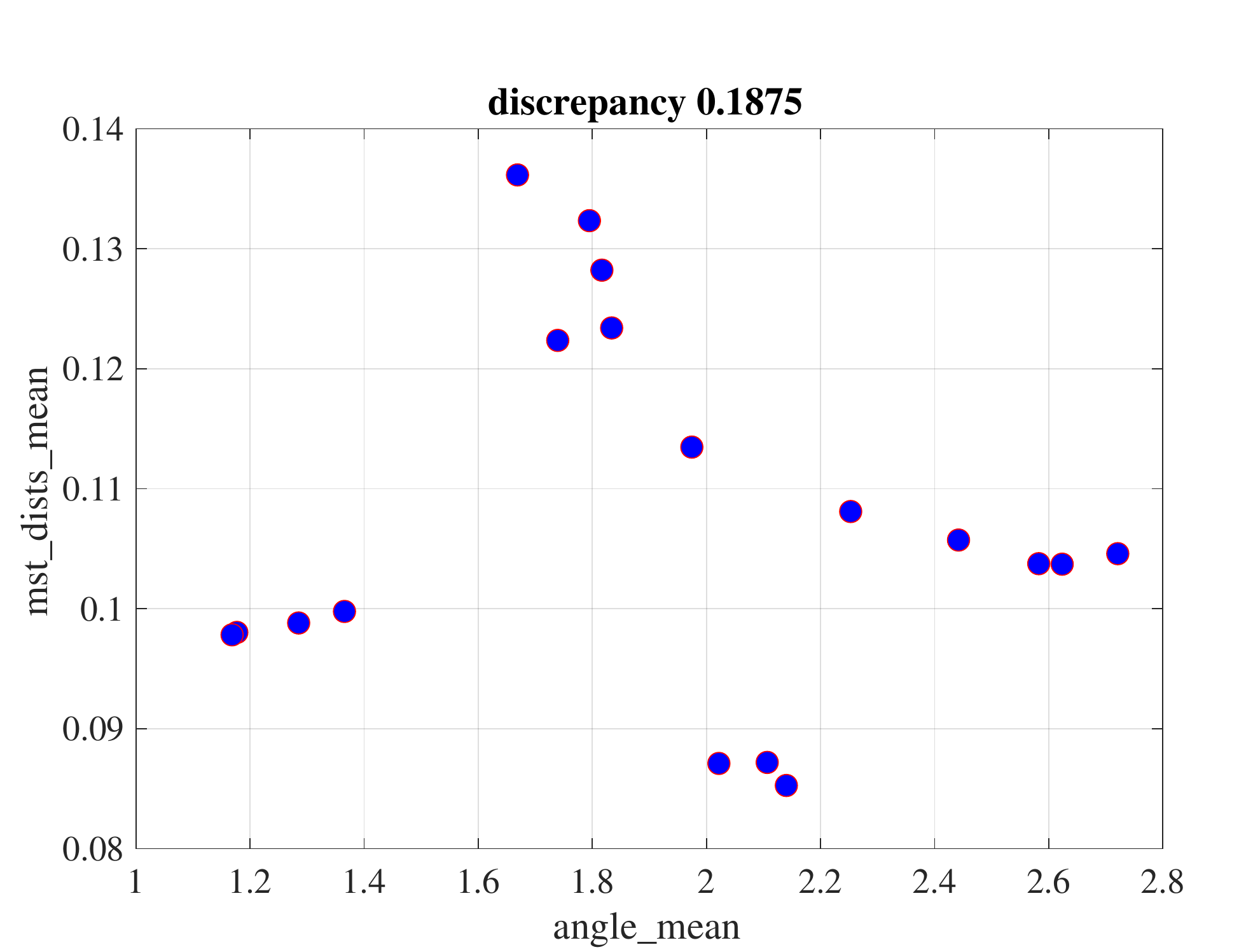}
\includegraphics[trim={0.5cm 7.7cm 0.7cm 6.8cm},clip,width=0.3\textwidth]{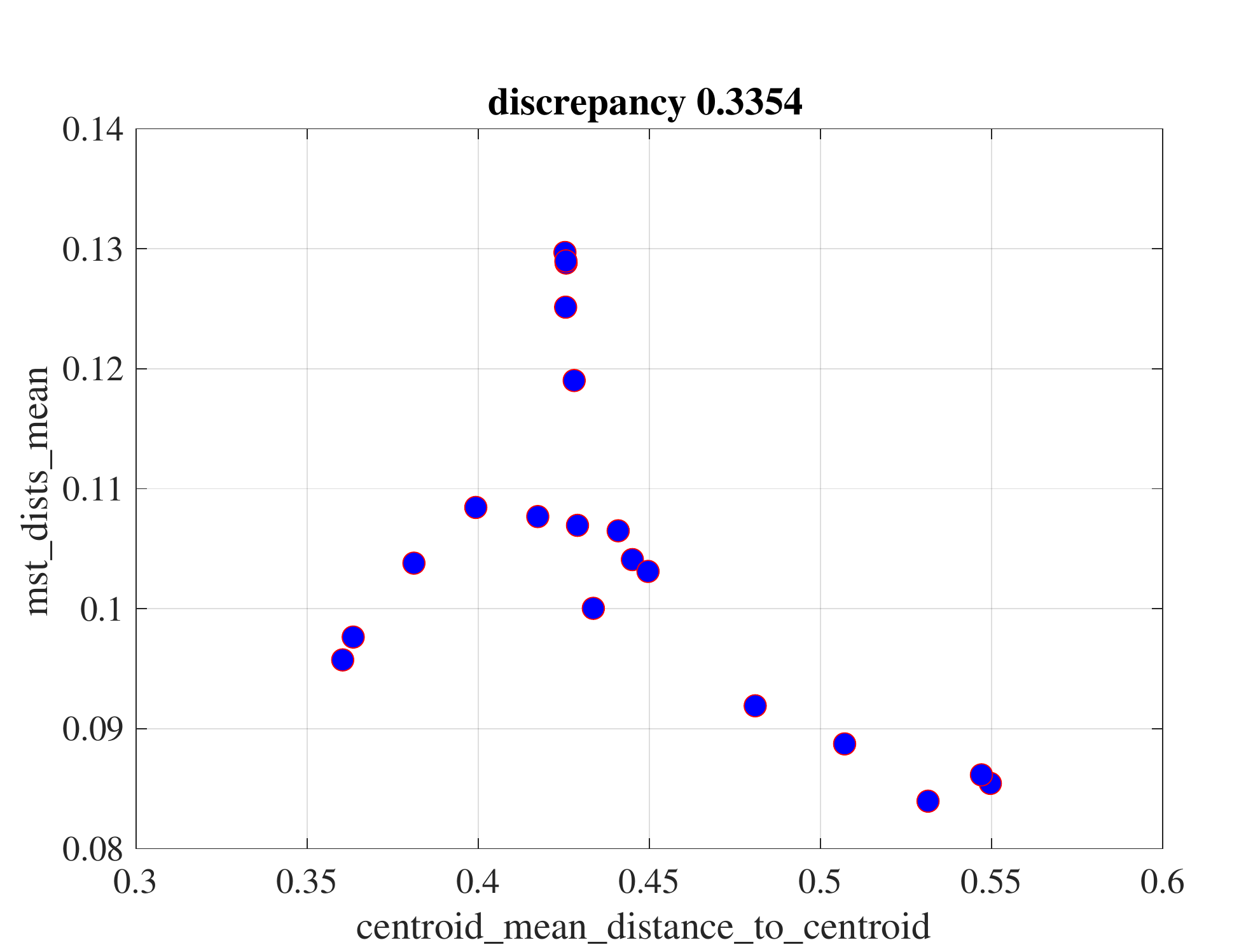}
\includegraphics[trim={0.5cm 7.7cm 0.7cm 6.8cm},clip,width=0.3\textwidth]{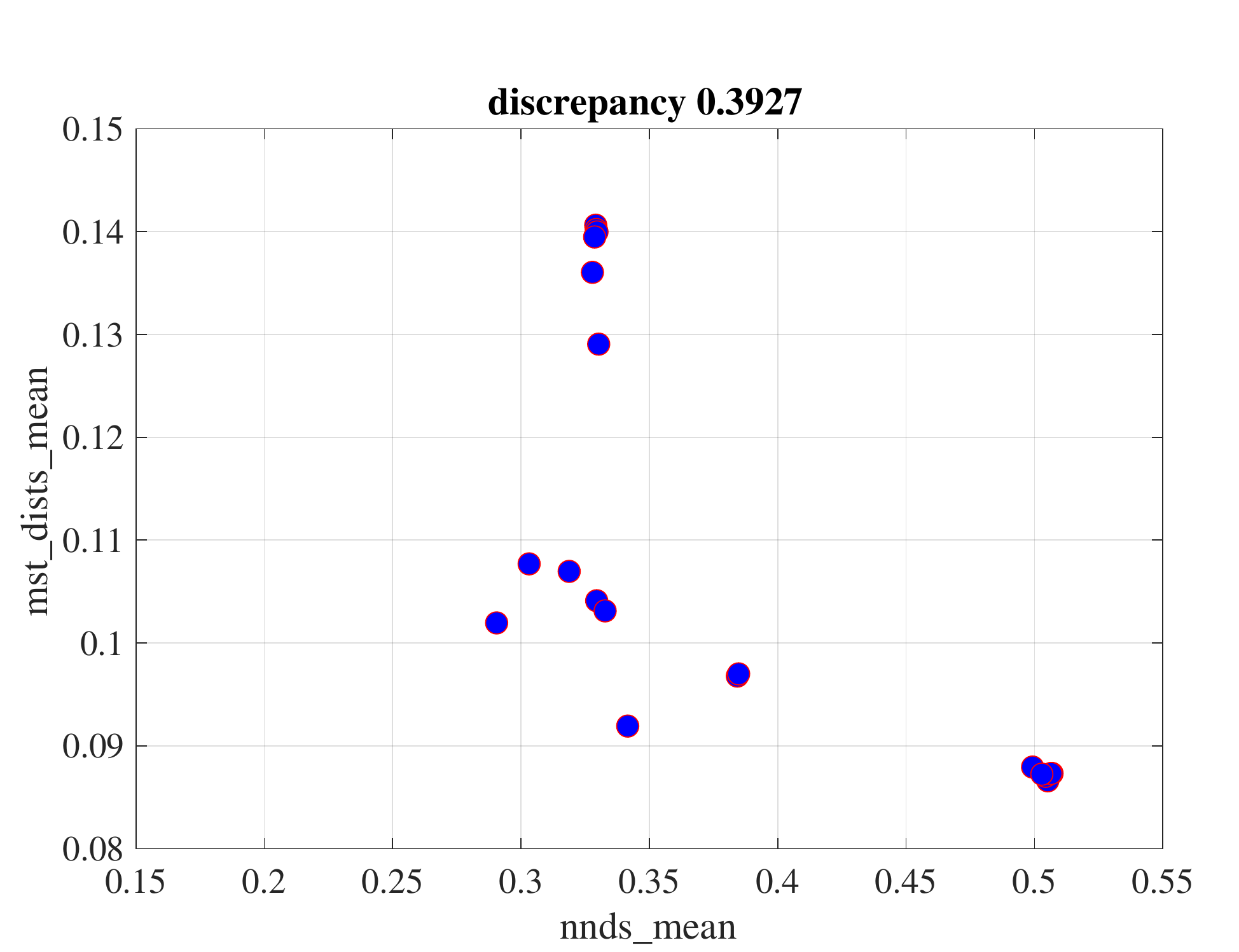}

\caption{Feature vectors for final population of $(\mu+\lambda)$-EA$_{C}$ (top) and  $(\mu+\lambda)$-EA$_{D}$ (bottom) for TSP based on two feature from left to right: (angle\_mean, mst\_dists\_mean), (centroid\_mean\_distance\_to\_centroid, mst\_dists\_mean), (nnds\_mean, mst\_dists\_mean) 
}.

\label{fig:tsp-2d}
\end{figure*}

Different combinations of features are tested in this research. The algorithms are designed to work with multiple features. As experiment, we choose three two-feature combination and three three-feature combination which are found to be better indicator of problem hardness.

All three algorithms are implemented in R and run in R environment~\cite{rManual}. We use the functions in tspmeta package to compute the feature values~\cite{Mersmann2013}.
All of the experiments are executed on a machine with $48$-core AMD $2.80$\,GHz CPU and $128$\,GByte RAM. 

Each algorithm is run for 20\,000 generations and the final discrepancy is reported. In order to obtain statistics, each feature combination is tested with each algorithm for 30 times. These 30 runs are independent to each other. 

\subsection{Experimental results and analysis}

Figure~\ref{fig:tsp-2d} shows the final population of TSP instances from the run which gets the minimum discrepancy value out of the 30 runs after applying Algorithm $(\mu+\lambda)$-$EA_{D}$ and $(\mu+\lambda)$-$EA_{C}$ in the feature space. 
The average initial discrepancy values for each feature combination in Table~\ref{tb:tsp} are $0.5786, 0.6090, 0.7227, 0.7997, 0.8142$ and $0.7699.$, respectively. 

The bottom row of  Figure~\ref{fig:tsp-2d} shows the feature vectors for final population of $(\mu+\lambda)$-$EA_{D}$. Compared to their counterparts in the bottom row, the discrepancy minimization approach generates a more diverse set for the feature combination of angle\_mean and mst\_dist\_mean. 
For the feature combination shown in the middle and on the right, it is not so obvious which algorithm generate a more diverse population than the other in the feature space. Each approach obtains a population that explores more over one feature value. For example, $(\mu+\lambda)$-$EA_{D}$ generates a population more diverse with respect to the feature of mst\_dists\_mean, while $(\mu+\lambda)$-EA$_{C}$ focuses more on exploring the feature space of centroid\_mean\_distance\_to\_centroid. Looking at the discrepancy values, it can be observed that the final population obtained by $(\mu+\lambda)$-$EA_{D}$ has a significantly smaller discrepancy than the one obtained by $(\mu+\lambda)$-EA$_{C}$ for all $3$ pairs of features.

Table ~\ref{tb:tsp} shows the statistics about the discrepancy values of the final populations after running each of the three algorithms on three 2-feature combinations and three 3-feature combinations.

The first two large columns contains the statistical results from $(\mu+\lambda)$-$EA_{C}$ and $(\mu+\lambda)$-$EA_{D}$. $(\mu+\lambda)$-$EA_{D}$ significantly outperforms $(\mu+\lambda)$-$EA_{C}$ in all feature combinations. The discrepancy value is reduced by more than 50\% in all six cases. Especially for the centroid\_mean\_dist\_centroid and mst\_dists\_mean feature combination, where $(\mu+\lambda)$-$EA_{D}$ achieves an average discrepancy value of 0.2268, which is less than half of the value obtained from $(\mu+\lambda)$-$EA_{C}$.

During the discrepancy minimization process, there exist many individuals which have the same least contribution to the discrepancy value in each iteration. Breaking ties according to the weighted feature contribution can help to improve the discrepancy of the population.
$(\mu+\lambda)$-$EA_{T}$ provides breaking ties with respect to the contribution to the weighting population diversity. The third column in Table~\ref{tb:tsp} shows the respective statistics for $(\mu+\lambda)$-$EA_{T}$. 
For the statistics, it shows $(\mu+\lambda)$-$EA_{T}$ is able to improve the discrepancy values of the final population. In all six examined feature combinations, $(\mu+\lambda)$-$EA_{T}$ achieves smaller discrepancy values than $(\mu+\lambda)$-$EA_{D}$.

\section{Conclusions}

Constructing point sets of low discrepancy has a prominent role in mathematics and a set of low discrepancy can be seen as being one that is covering the considered space $[0,1]^d$ in a good way as they aim for a good balance of points in every hyper-box with respect to their volume. We have introduced a discrepancy-based evolutionary diversity optimization approach which constructs sets of solution meeting a given quality criteria and having a low discrepancy with respect to the considered features. Our experimental results for evolving diverse sets of images and TSP instances show that this approach constructs sets of solutions with a much lower discrepancy that the previously used weighted contribution approach according to the given features. 
Our discrepancy-based diversity optimization process for images makes use of a new random walk mutation operator which reduces the number of required generations to obtain a good diverse set of images by 3 orders of magnitude
The best results across all our experimental investigations are obtained by $(\mu+\lambda)$-EA$_{T}$ which uses discrepancy-based diversity optimization in conjunction with a tie-breaking rule based on the weighted contribution diversity measure.

\begin{sidewaystable}
\centering
\begin{tabular}{lccccccccc}
 & \multicolumn{3}{c}{$(\mu+\lambda)$-$EA_{C}$}
 & \multicolumn{3}{c}{$(\mu+\lambda)$-$EA_{D}$}  &  \multicolumn{3}{c}{$(\mu+\lambda)$-$EA_{T}$} \\ 
 & min & mean & std  & min & mean & std &   min & mean & std \\ \cmidrule(l){2-10}

\multicolumn{1}{l|}{( SDHue, Saturation )} & 
0.2014  &  0.3234 &  \multicolumn{1}{l|}{0.0595} & 0.1272 & 0.2038 & \multicolumn{1}{l|}{0.1157} &  0.1119 &  0.1530 & 0.0269  \\

\multicolumn{1}{l|}{( Symmetry, Hue )} & 
 0.1964 &  0.2945 &  \multicolumn{1}{l|}{0.0497} & 0.1574 & 0.2280 & \multicolumn{1}{l|}{0.0592} &  0.1051 &  0.1417  & 0.0179  \\
 
 \multicolumn{1}{l|}{( GCF, Smoothness )} & 
0.1997 & 0.2769 &  \multicolumn{1}{l|}{0.0344} & 0.1363 & 0.2025 & \multicolumn{1}{l|}{0.0538}  & 0.1457 &  0.1800  & 0.0234 \\

 \multicolumn{1}{l|}{( SDHue, Saturation, Symmetry )}
&0.3389 &  0.4327  & \multicolumn{1}{l|}{0.0613 }  &   0.1513 & 0.3335 &  \multicolumn{1}{l|} { 0.1062} &  0.2253 &  0.2814 & 0.0422  \\
 
\multicolumn{1}{l|}{( SDHue, Hue, Symmetry )} & 0.2754 &  0.3395  &  \multicolumn{1}{l|}{0.0483} & 0.2100 & 0.3118 & \multicolumn{1}{l|}{0.1309 }
 & 0.2224 & 0.2600 & 0.0123  \\
 
\multicolumn{1}{l|}{( GCF, Hue, Saturation )} &  0.4775  &  0.6488 &   \multicolumn{1}{l|}{0.0841} &   0.2021  &  0.3007 & \multicolumn{1}{l|}{ 0.1467} &  0.1983 & 0.2229 &  0.0125  \\ 
 
\end{tabular}\vspace{1mm}
\caption{Statistics of discrepancy values for Images.}
\label{tb:images}

\vspace{15mm}

\begin{tabular}{@{}lllllllllllll@{}}
 &  \multicolumn{3}{c}{$(\mu+\lambda)$-$EA_{C}$} &  \multicolumn{3}{c}{$(\mu+\lambda)$-$EA_{D}$} &  \multicolumn{3}{c}{$(\mu+\lambda)$-$EA_{T}$} \\ 
 & min & mean & std  & min & mean & std & min & mean & std \\ \cmidrule(l){2-10} 
 \multicolumn{1}{l|}{(angle\_mean,mst\_dists\_mean)} & 0.4836 & 0.5522 &  \multicolumn{1}{l|}{0.0361} &  0.1875 & 0.3059 & \multicolumn{1}{l|}{0.1192} & 0.2043 & 0.2456 & 0.0312 \\
 \multicolumn{1}{l|}{(centroid\_mean\_dist\_centroid, mst\_dists\_mean)} & 0.4657 &  0.5522 & \multicolumn{1}{l|}{0.0361} & 0.3354 & 0.3777 & \multicolumn{1}{l|}{0.0490} & 0.3088 & 0.3377 & 0.0182 \\
 \multicolumn{1}{l|}{(nnds\_mean,mst\_dists\_mean)} & 0.5743 & 0.6296 & \multicolumn{1}{l|}{0.0219} & 0.3927 & 0.4422 & \multicolumn{1}{l|}{0.0534} & 0.3831 & 0.4107 & 0.0193 \\
 \multicolumn{1}{l|}{(angle\_mean,nnds\_mean,mst\_dists\_mean)} & 0.7838 & 0.8005 & \multicolumn{1}{l|}{0.0244}  & 0.4354 & 0.4547 & \multicolumn{1}{l|}{0.0534} &  0.4322 & 0.4451 & 0.0174 \\
 \multicolumn{1}{l|}{{\small(centroid\_mean\_dist\_centroid,nnds\_mean,mst\_dists\_mean)}} & 0.7641 & 0.7895 & \multicolumn{1}{l|}{0.0155} &  0.4197 & 0.4547 & \multicolumn{1}{l|}{0.0191} &  0.3730 & 0.4542 & 0.0346 \\
 \multicolumn{1}{l|}{{\small(angle\_mean,centroid\_mean\_dist\_centroid,nnds\_mean)}} & 0.7593 & 0.7809 & \multicolumn{1}{l|}{0.0131} &  0.3912 & 0.4095 & \multicolumn{1}{l|}{0.0135} & 0.3547 & 0.3986 & 0.0257 \\ 
\end{tabular}\vspace{1mm}
\caption{Statistics of discrepancy values for TSP.}
\label{tb:tsp}
\end{sidewaystable}



\end{document}